\documentclass[11pt,a4paper]{article}

\usepackage{amsmath, amssymb, mathrsfs}
\usepackage{latexsym, color}
\usepackage{pgf,tikz}
\usetikzlibrary{arrows}

 \usepackage{ulem}

\usepackage{epsfig}

\usepackage{hyperref}

\numberwithin{equation}{section}

\begin{document}

\newtheorem{thm}{Theorem}[section]
\newtheorem{cor}[thm]{Corollary}
\newtheorem{lem}[thm]{Lemma}
\newtheorem{prop}[thm]{Proposition}
\newtheorem{defn}[thm]{Definition} 
\newtheorem{rem}[thm]{Remark}
\newtheorem{Ex}[thm]{EXAMPLE}
\def\nm{\noalign{\medskip}}

\bibliographystyle{plain}


\newcommand{\qed}{\hfill \ensuremath{\square}}
\newcommand{\ds}{\displaystyle}
\newcommand{\pf}{\noindent {\sl Proof}. \ }
\newcommand{\p}{\partial}
\newcommand{\pd}[2]{\frac {\p #1}{\p #2}}
\newcommand{\norm}[1]{\left\| #1 \right\|}
\newcommand{\dbar}{\overline \p}
\newcommand{\eqnref}[1]{(\ref {#1})}
\newcommand{\na}{\nabla}
\newcommand{\one}[1]{#1^{(1)}}
\newcommand{\two}[1]{#1^{(2)}}
\newcommand{\proj}{\mbox{\normalfont proj}}

\newcommand{\Abb}{\mathbb{A}}
\newcommand{\Cbb}{\mathbb{C}}
\newcommand{\Ibb}{\mathbb{I}}
\newcommand{\Nbb}{\mathbb{N}}
\newcommand{\Kbb}{\mathbb{K}}
\newcommand{\Rbb}{\mathbb{R}}
\newcommand{\Sbb}{\mathbb{S}}

\renewcommand{\div}{\mbox{div}~}

\newcommand{\la}{\langle}
\newcommand{\ra}{\rangle}
\newcommand{\Ecal}{\mathcal{E}}
\newcommand{\Fcal}{\mathcal{F}}
\newcommand{\Hcal}{\mathcal{H}}
\newcommand{\Lcal}{\mathcal{L}}
\newcommand{\Kcal}{\mathcal{K}}
\newcommand{\Dcal}{\mathcal{D}}
\newcommand{\Pcal}{\mathcal{P}}
\newcommand{\Qcal}{\mathcal{Q}}
\newcommand{\Scal}{\mathcal{S}}
\newcommand{\Ncal}{\mathcal{N}}

\def\Ba{{\bf a}}
\def\Bb{{\bf b}}
\def\Bc{{\bf c}}
\def\Bd{{\bf d}}
\def\Be{{\bf e}}
\def\Bf{{\bf f}}
\def\Bg{{\bf g}}
\def\Bh{{\bf h}}
\def\Bi{{\bf i}}
\def\Bj{{\bf j}}
\def\Bk{{\bf k}}
\def\Bl{{\bf l}}
\def\Bm{{\bf m}}
\def\Bn{{\bf n}}
\def\Bo{{\bf o}}
\def\Bp{{\bf p}}
\def\Bq{{\bf q}}
\def\Br{{\bf r}}
\def\Bs{{\bf s}}
\def\Bt{{\bf t}}
\def\Bu{{\bf u}}
\def\Bv{{\bf v}}
\def\Bw{{\bf w}}
\def\Bx{{\bf x}}
\def\By{{\bf y}}
\def\Bz{{\bf z}}
\def\BA{{\bf A}}
\def\BB{{\bf B}}
\def\BC{{\bf C}}
\def\BD{{\bf D}}
\def\BE{{\bf E}}
\def\BF{{\bf F}}
\def\BG{{\bf G}}
\def\BH{{\bf H}}
\def\BI{{\bf I}}
\def\BJ{{\bf J}}
\def\BK{{\bf K}}
\def\BL{{\bf L}}
\def\BM{{\bf M}}
\def\BN{{\bf N}}
\def\BO{{\bf O}}
\def\BP{{\bf P}}
\def\BQ{{\bf Q}}
\def\BR{{\bf R}}
\def\BS{{\bf S}}
\def\BT{{\bf T}}
\def\BU{{\bf U}}
\def\BV{{\bf V}}
\def\BW{{\bf W}}
\def\BX{{\bf X}}
\def\BY{{\bf Y}}
\def\BZ{{\bf Z}}


\newcommand{\Ga}{\alpha}
\newcommand{\Gb}{\beta}
\newcommand{\Gd}{\delta}
\newcommand{\Ge}{\epsilon}
\newcommand{\Gve}{\varepsilon}
\newcommand{\Gf}{\phi}
\newcommand{\Gvf}{\varphi}
\newcommand{\Gg}{\gamma}
\newcommand{\Gc}{\chi}
\newcommand{\Gi}{\iota}
\newcommand{\Gk}{\kappa}
\newcommand{\Gvk}{\varkappa}
\newcommand{\Gl}{\lambda}
\newcommand{\Gn}{\eta}
\newcommand{\Gm}{\mu}
\newcommand{\Gv}{\nu}
\newcommand{\Gp}{\pi}
\newcommand{\Gt}{\theta}
\newcommand{\Gvt}{\vartheta}
\newcommand{\Gr}{\rho}
\newcommand{\Gvr}{\varrho}
\newcommand{\Gs}{\sigma}
\newcommand{\Gvs}{\varsigma}
\newcommand{\Gj}{\tau}
\newcommand{\Gu}{\upsilon}
\newcommand{\Go}{\omega}
\newcommand{\Gx}{\xi}
\newcommand{\Gy}{\Psi}
\newcommand{\Gz}{\zeta}
\newcommand{\GD}{\Delta}
\newcommand{\GF}{\Phi}
\newcommand{\GG}{\Gamma}
\newcommand{\GL}{\Lambda}
\newcommand{\GP}{\Pi}
\newcommand{\GT}{\Theta}
\newcommand{\GS}{\Sigma}
\newcommand{\GU}{\Upsilon}
\newcommand{\GO}{\Omega}
\newcommand{\GX}{\Xi}
\newcommand{\GY}{\Psi}
\newcommand{\idx}{\mathscr{I}}
\newcommand{\idxp}{\mathscr{I}_+}
\newcommand{\beq}{\begin{equation}}
\newcommand{\eeq}{\end{equation}}

\title{ An explicit  formulation of  the learned noise predictor $\epsilon_{\theta}({\bf x}_t, t)$ via  the forward-process noise  $\epsilon_{t}$\\ in denoising diffusion probabilistic models (DDPMs)}

\author{KiHyun Yun \thanks{\footnotesize
Department of Mathematics, Hankuk University of Foreign Studies,
Youngin-si, Gyeonggi-do 449-791, Republic of Korea
(kihyun.yun@gmail.com).}}

\maketitle

\date{}

 \maketitle
 
 \abstract { In denoising diffusion probabilistic models (DDPMs), the learned noise predictor $ \epsilon_{\theta} ( {\bf x}_t ,  t)$ is trained to approximate the forward-process noise $\epsilon_t$. The equality $\nabla_{{\bf x}_t} \log q({\bf x}_t)  = -\frac 1 {\sqrt {1- {\bar \alpha}_t} } \epsilon_{\theta} ( {\bf x}_t ,  t)$ plays a fundamental role in both theoretical analyses and algorithmic design, and thus is frequently employed across diffusion-based generative models.  In this paper,  an explicit formulation of $ \epsilon_{\theta} ( {\bf x}_t ,  t)$ in terms of the forward-process noise $\epsilon_t$ is derived. This result show how the forward-process noise $\epsilon_t$ contributes to the learned predictor $ \epsilon_{\theta} ( {\bf x}_t ,  t)$. Furthermore,  based on this formulation, we present a novel and mathematically rigorous proof of the fundamental equality above, clarifying its origin and providing new theoretical insight into the structure of diffusion models.}

\section { Introduction} 
 Denoising diffusion probabilistic models (DDPMs) is a basic framework for generative modeling, wherein a neural network $ \epsilon_{\theta} ( {\bf x}_t ,  t)$ is trained to predict the noise $\epsilon_t$ added in a forward diffusion process. In this paper, we use the definitions and notations as presented in DDPM paper \cite{Ho}. A fundamental identity that appears repeatedly in diffusion-based generative models is
\beq \nabla_{{\bf x}_t} \log q({\bf x}_t)  = -\frac 1 {\sqrt {1- {\bar \alpha}_t} } \epsilon_{\theta} ( {\bf x}_t ,  t) \label{fundam_equality}\eeq
which connects the score function of the marginal distribution $q({\bf x}_t)$ to the learned noise predictor $ \epsilon_{\theta} ( {\bf x}_t ,  t)$. Refer to \cite{DN, Ho2, Ni}. This identity can be derived via Tweedie’s formula, or alternatively as a byproduct of the reverse-time SDE framework established by Anderson \cite{And}. Also refer to \cite{Song}. However, both derivations do not clearly show how the forward-process noise $\epsilon_t$ contributes to the learned predictor $ \epsilon_{\theta} ( {\bf x}_t ,  t)$, and why it ends up matching the score function of the marginal distribution $q({\bf x}_t)$ precisely.

\par In this work, we investigate this connection and provide an explicit formulation that expresses $ \epsilon_{\theta} ( {\bf x}_t ,  t)$ as the conditional expectation of the forward-process noise  $\epsilon_t$ under the posterior $q({\bf x}_0|{\bf x}_t)$. This perspective not only yields a novel and mathematically rigorous proof of the identity above, but also offers a clearer understanding of how the denoising objective implicitly captures the score of the marginal distribution. Our result describes the mathematical structure underlying DDPMs and reinforces the interpretation of noise prediction as a form of conditional averaging.

\par We begin by clarifying the definition of $ \epsilon_{\theta} ( {\bf x}_t ,  t)$. We use the definitions and notations as presented in DDPM paper \cite{Ho}, and thus refer to the paper \cite{Ho} for details. Let 
${\bf x}_0$  be a sample from the data distribution $q({\bf x}_0)$. The forward diffusion process generates a sequence of latent variables 
${\bf x}_1,~\cdots, {\bf x}_T$ by gradually adding Gaussian noise. For each timestep $t = 1, \cdots, T$, the variable ${\bf x}_t$ in defined by  
$${\bf x}_t = \sqrt { {\bar \alpha}_t}  {\bf x}_0 + \sqrt {1- {\bar \alpha}_t} \epsilon_t, ~\mbox{where}~ \epsilon_t \sim {\Ncal}({\bf 0}, \mathbb{I}).$$ Here, $ {\bar \alpha}_t \in (0,1)$ is the cumulative product of the noise schedule. It is worthy to note that $\epsilon_t$ is given by ${\bf x}_0 $ and ${\bf x}_t$. Thus, we define a formal notation $ \epsilon_t( {\bf x}_t |  {\bf x}_0 )$ for $ \epsilon_t $ above as
$$ \epsilon_t( {\bf x}_t |  {\bf x}_0 ) =  \epsilon_t.$$
From the definition of $\epsilon_{\theta} (\cdot,t)$ in DDPM, we find the parameter $\tilde{\theta}$ minimizing the expection
\begin{align*} 
&\mathbb{E}_{{\bf x }_0 \sim q ({\bf x }_0)} \left [\mathbb{E}_{ \epsilon_t   \sim {\Ncal}({\bf 0}, \mathbb{I}) } \left[ \norm{\epsilon_t  - \epsilon_{\tilde\theta} \left( \sqrt { {\bar \alpha}_t}  {\bf x}_0 + \sqrt {1- {\bar \alpha}_t} \epsilon_t, t \right)  }^2\right] \right].
\end{align*}
We define the function $\epsilon_{\theta} (\cdot,t)$ of ${\bf x}_t$ as
\begin{align*} 
 & \epsilon_{\theta} (\cdot,t)= f_* (\cdot)\\&= \mbox{argmin}_{f (\cdot)}   \mathbb{E}_{{\bf x }_0 \sim q ({\bf x }_0)}  \left [\mathbb{E}_{ \epsilon_t  \sim {\Ncal} ({\bf 0}, \mathbb{I}) } \left[ \norm{\epsilon_t  - f \left(  \sqrt { {\bar \alpha}_t}  {\bf x}_0 + \sqrt {1- {\bar \alpha}_t} \epsilon_t \right)  }^2 \right] \right].
\end{align*}  Please note that the above 
$f_*$ is not used commonly regardless of the choice of $t$. There is a different function $f_*$ for each $t$.

\begin{thm}\label{main} Let $\epsilon_{\theta} (\cdot,t)$ and $ \epsilon_t( {\bf x}_t |  {\bf x}_0 )$ be as given above. Then, we have 
$$\epsilon_{\theta} ({\bf x}_t,t) = \int_{\Omega({\bf X}_0)}  \epsilon_t( {\bf x}_t |  {\bf x}_0 ) q({\bf x}_0 | {\bf x}_t) d{\bf x}_0. $$ where $\Omega({\bf X}_0)$ is the domain of ${\bf X}_0$.
\end{thm}
The proof is presented in Section 2.  The representation of $\epsilon_{\theta} ({\bf x}_t,t)$ is reasonable by noting that, since $\epsilon_{\theta} ({\bf x}_t,t)$ is a function of ${\bf x}_t$, it should be considered under the condition with  ${\bf x}_t$ is given, and consequently, the distribution of $q({\bf x}_0 | {\bf x}_t)$ is used to describe $\epsilon_{\theta} ({\bf x}_t,t)$.

\par The preceding theorem provides a natural derivation of the fundamental identity \eqref{fundam_equality}, which is also stated in the following corollary.

\begin{cor}\label{cor}
$$\nabla_{{\bf x}_t} \log q({\bf x}_t)  = -\frac 1 {\sqrt {1- {\bar \alpha}_t} } \epsilon_{\theta} ( {\bf x}_t ,  t).$$
\end{cor}
The equality is found in (11) in \cite{DN}. The following provides a new derivation the equality above derived from Theorem \ref{main}.
\vskip 10pt
\par {\noindent {\sl Proof of Corollary} \ref{cor}. \ } We begin by considering 
$$ \epsilon_t( {\bf x}_t |  {\bf x}_0 ) =  \epsilon_t, ~\mbox{ and } {\bf x}_t = \sqrt { {\bar \alpha}_t}  {\bf x}_0 + \sqrt {1- {\bar \alpha}_t} \epsilon_t,$$ where $\epsilon_t \sim {\Ncal}({\bf 0}, \mathbb{I}).$ Then we have
\begin{align}
 \frac { {\bf x}_0 -\sqrt{ {\bar \alpha}_t}    { {\bf x}_0}    }   {\sqrt {1- { {\bar \alpha}_t}  } } 
&=\epsilon_t( {\bf x}_t |  {\bf x}_0 ) \sim {\Ncal}({\bf 0}, \mathbb{I}), \label{step1-1}\\
q({\bf x }_t | {\bf x }_0) &= p(\epsilon_t( {\bf x}_t |  {\bf x}_0 ) | {\bf x}_0 ) = {\Ncal}( \epsilon_t( {\bf x}_t |  {\bf x}_0 ) | {\bf 0}, \mathbb{I}). \label{step1-2} \end{align}
In details,
\begin{align*}
q({\bf x }_t | {\bf x }_0) &=  {\Ncal}({\bf x}_t | \sqrt { {\bar \alpha}_t} { {\bf x}_0}, (1- {\bar \alpha}_t)\mathbb{I})  = \frac 1 Z \exp \left( - {\frac {1} { 2 (1- {\bar \alpha}_t) }} \norm {{\bf x}_t - \sqrt { {\bar \alpha}_t} { {\bf x}_0}  }^2\right).
\end{align*} Then, by the relation \eqref{step1-1},
\begin{align}
\nabla_{{\bf x }_t} \log q({\bf x }_t | {\bf x }_0) &=  - {\frac {1} { 1- {\bar \alpha}_t }} \left({\bf x}_t - \sqrt { {\bar \alpha}_t} { {\bf x}_0} \right) \notag\\
&= -  {\frac {1} { \sqrt {1- {\bar \alpha}_t }}}\epsilon_t( {\bf x}_t |  {\bf x}_0 ).\label{grad_log_q}
\end{align}

\par Applying \eqref {grad_log_q} to the result of Theorem \ref{main}, 
\begin{align*}   \epsilon_{\theta} ({\bf x}_t,t) q({\bf x}_t ) &=  \left(\int_{\Omega{({\bf x}_0)} } \epsilon_t ( {\bf x}_t |  {\bf x}_0 ) q({\bf x}_0 | {\bf x}_t)  d{\bf x}_0 \right)  q({\bf x}_t ) \\ &=  \int_{\Omega{({\bf x}_0)} } \epsilon_t ( {\bf x}_t |  {\bf x}_0 ) q({\bf x}_t , {\bf x}_0)  d{\bf x}_0 \\
&=  \int_{\Omega{({\bf x}_0)} } - { \sqrt {1- {\bar \alpha}_t }}  \Big(\nabla_{{\bf x }_t} \log q({\bf x }_t | {\bf x }_0)  \Big) q({\bf x}_t | {\bf x}_0) q( {\bf x}_0) d{\bf x}_0 \\
&=  - { \sqrt {1- {\bar \alpha}_t }}  \int_{\Omega{({\bf x}_0)} }  \left(\frac {\nabla_{{\bf x }_t}  q({\bf x }_t | {\bf x }_0)}{q({\bf x }_t | {\bf x }_0)}  \right) q({\bf x}_t | {\bf x}_0) q( {\bf x}_0) d{\bf x}_0 \\
&=  - { \sqrt {1- {\bar \alpha}_t }} \nabla_{{\bf x }_t}  \int_{\Omega{({\bf x}_0)} }  \left(\frac {  q({\bf x }_t , {\bf x }_0)}{q( {\bf x }_0)}  \right) q( {\bf x}_0) d{\bf x}_0 \\
&=  - { \sqrt {1- {\bar \alpha}_t }} \nabla_{{\bf x }_t}  \int_{\Omega{({\bf x}_0)} }  {  q({\bf x }_t , {\bf x }_0)} d{\bf x}_0 \\
&=  - { \sqrt {1- {\bar \alpha}_t }} \nabla_{{\bf x }_t}   q({\bf x }_t).
 \end{align*}
Then, $$ \epsilon_{\theta} ({\bf x}_t,t)    =  - { \sqrt {1- {\bar \alpha}_t }} \frac {\nabla_{{\bf x }_t}  q({\bf x }_t) }{ q({\bf x }_t)} = - { \sqrt {1- {\bar \alpha}_t }}  \nabla_{{\bf x }_t} \log  q({\bf x }_t) .$$ Thus, we have this corollary. \qed

\section { Proof of the theorem \ref{main}}
\par Although a fully rigorous proof would require explicitly specifying the appropriate function space and proceeding accordingly, we have chosen instead to present a concise proof that reveals the core idea behind the argument. This approach is intended to make the content more accessible to a broader readers, including readers without a background in mathematics.

\par From the definition of $\epsilon_t ( {\bf x}_t |  {\bf x}_0 )$, the expectation to be minimized for defining the function $\epsilon_{\theta} ({\bf x}_t,t) $ can be reformulated and understood in the following form:
\begin{align*} 
&\mathbb{E}_{{\bf x }_0 \sim q ({\bf x }_0)} \left [\mathbb{E}_{ \epsilon_t  = \epsilon_t ( {\bf x}_t |  {\bf x}_0 ) \sim {\Ncal}({\bf 0}, \mathbb{I}) } \left[ \norm{\epsilon_t  - \epsilon_{\theta} \left( \sqrt { {\bar \alpha}_t}  {\bf x}_0 + \sqrt {1- {\bar \alpha}_t} \epsilon_t, t \right)  }^2\right] \right] \\ 
&= \mathbb{E}_{{\bf x }_0 \sim q ({\bf x }_0)} \left [\mathbb{E}_{ \epsilon_t  \sim {\Ncal} ({\bf 0}, \mathbb{I}) } \left[ \norm{\epsilon_t  - \epsilon_{\theta} \left( {\bf x}_t , t \right)  }^2 ~\big|~  {\bf x}_t = \sqrt { {\bar \alpha}_t}  {\bf x}_0 + \sqrt {1- {\bar \alpha}_t} \epsilon_t \right] \right],
\end{align*} and the function $\epsilon_{\theta} \left( \cdot, t \right)$ of ${\bf x}_t$ is a function  $f_* (\cdot)$ minimizing the expectation as follows:
\begin{align} 
&\epsilon_{\theta} \left( \cdot, t \right)  = f_* (\cdot) \label{f_*} \\
&= \mbox{argmin}_{f (\cdot)} \notag \\& \mathbb{E}_{{\bf x }_0 \sim q ({\bf x }_0)} \left [\mathbb{E}_{\epsilon_t ( {\bf x}_t |  {\bf x}_0 ) \sim {\Ncal} ({\bf 0}, \mathbb{I}) } \left[ \norm{\epsilon_t ( {\bf x}_t |  {\bf x}_0 )   - f \left( {\bf x}_t \right)  }^2 ~\big|~  {\bf x}_t = \sqrt { {\bar \alpha}_t}  {\bf x}_0 + \sqrt {1- {\bar \alpha}_t}\epsilon_t ( {\bf x}_t |  {\bf x}_0 ) \right] \right]\notag\\
&= \mbox{argmin}_{f (\cdot)} \mathbb{E}_{{\bf x }_0 \sim q ({\bf x }_0)} \left [\mathbb{E}_{  {\bf x}_t \sim q ( {\bf x}_t |  {\bf x}_0 )  } \left[ \norm{\epsilon_t ( {\bf x}_t |  {\bf x}_0 )   - f \left( {\bf x}_t \right)  }^2  \right] \right].\notag
\end{align}  Let $h ( \cdot)$ be arbitrary function, and $s$ be a real number. Then, 
$$ (f_* + s h) ({\bf x}_t) =  f_* ({\bf x}_t) +  s h ({\bf x}_t).$$  At $s=0$, the function $(f_* + s h) ({\bf x}_t)$ minimizes the following expection
$$F_h (s) =  \mathbb{E}_{{\bf x }_0 \sim q ({\bf x }_0)} \left [\mathbb{E}_{  {\bf x}_t \sim q ( {\bf x}_t |  {\bf x}_0 )  } \left[ \norm{\epsilon_t ( {\bf x}_t |  {\bf x}_0 )   - (f_* + s h)  \left( {\bf x}_t \right)  }^2  \right] \right].$$ Let $\Omega({\bf x}_t)$ be the set of all ${\bf x}_t$, and it is the common domain of $f_*$ and $h$. 
Then, \begin{align} 
0&=\frac {d F_h }{d s} (0) \notag
\\&=\mathbb{E}_{{\bf x }_0 \sim q ({\bf x }_0)} \left [\mathbb{E}_{  {\bf x}_t \sim q ( {\bf x}_t |  {\bf x}_0 )  } \Big[ 2 \left(\epsilon_t ( {\bf x}_t |  {\bf x}_0 )   - f_* \left( {\bf x}_t \right)  \right) h({\bf x}_t ) \Big] \right] \notag\\
&= 2 \int_{\Omega{({\bf x}_0)}}\left(\int_{\Omega_{({\bf x}_t)} }  \left(\epsilon_t ( {\bf x}_t |  {\bf x}_0 )   - f_* \left( {\bf x}_t \right)  \right) h({\bf x}_t ) q({\bf x}_t | {\bf x}_0) d {\bf x}_t \right)q ({\bf x}_0)d{\bf x}_0\notag\\
&= 2 \int_{\Omega{({\bf x}_t)}}\left(  \int_{\Omega_{({\bf x}_0)} }  \left(\epsilon_t ( {\bf x}_t |  {\bf x}_0 )   - f_* \left( {\bf x}_t \right)  \right)  q({\bf x}_t | {\bf x}_0)  q ({\bf x}_0)d{\bf x}_0\right) h({\bf x}_t ) d{\bf x}_t\notag\\
&= 2 \int_{\Omega{({\bf x}_t)}}g({\bf x}_t )  h({\bf x}_t ) d{\bf x}_t, \label{zero_equal}
\end{align} where
$$g({\bf x}_t ) =   \int_{\Omega{({\bf x}_0)} }  \left(\epsilon_t ( {\bf x}_t |  {\bf x}_0 )   - f_* \left( {\bf x}_t \right)  \right)  q({\bf x}_t | {\bf x}_0)  q ({\bf x}_0)d{\bf x}_0.$$ Note that $g$ is regardless of $h$.  Since  $h$ is an arbitrary function, the equality \eqref{zero_equal} implies
\begin{align*} 
0 &= g({\bf x}_t ) =   \int_{\Omega{({\bf x}_0)} }  \left(\epsilon_t ( {\bf x}_t |  {\bf x}_0 )   - f_* \left( {\bf x}_t \right)  \right)  q({\bf x}_t | {\bf x}_0)  q ({\bf x}_0)d{\bf x}_0 \\
&= \int_{\Omega{({\bf x}_0)} }  \left(\epsilon_t ( {\bf x}_t |  {\bf x}_0 )   - f_* \left( {\bf x}_t \right)  \right)  q({\bf x}_t , {\bf x}_0)  d{\bf x}_0\\
&= \int_{\Omega{({\bf x}_0)} } \epsilon_t ( {\bf x}_t |  {\bf x}_0 ) q({\bf x}_t , {\bf x}_0)  d{\bf x}_0    -   f_* \left( {\bf x}_t \right)  \int_{\Omega_{({\bf x}_0)} }   q({\bf x}_t , {\bf x}_0)  d{\bf x}_0 \\
&= \int_{\Omega{({\bf x}_0)} } \epsilon_t ( {\bf x}_t |  {\bf x}_0 ) q({\bf x}_t , {\bf x}_0)  d{\bf x}_0    -   f_* \left( {\bf x}_t \right)    q({\bf x}_t ).
\end{align*}
Then, \begin{align*}  
\epsilon_{\theta} ({\bf x}_t,t)  &= f_* \left( {\bf x}_t \right) \\&= \frac 1 {q({\bf x}_t )}  \int_{\Omega{({\bf x}_0)} } \epsilon_t ( {\bf x}_t |  {\bf x}_0 ) q({\bf x}_t , {\bf x}_0)  d{\bf x}_0\\& = \int_{\Omega{({\bf x}_0)} } \epsilon_t ( {\bf x}_t |  {\bf x}_0 ) q({\bf x}_0 |{\bf x}_t)  d{\bf x}_0 
\end{align*} Thus, we have the desirable result.\qed

\end{document}